\documentclass{article}

\usepackage[final]{nips_2018}

\usepackage[utf8]{inputenc} 
\usepackage[T1]{fontenc}    
\usepackage{hyperref}       
\usepackage{url}            
\usepackage{booktabs}       
\usepackage{amsfonts}       
\usepackage{nicefrac}       
\usepackage{microtype}      
\usepackage{graphicx}
\usepackage{amsmath,amsfonts,amsthm,bm, amssymb}
\usepackage{caption}
\usepackage{subcaption}
\usepackage{color}

\title{Transferring Physical Motion Between Domains for Neural Inertial Tracking}
\author{
  Changhao Chen$^{1}$, Yishu Miao$^{1}$, Chris Xiaoxuan Lu$^{1}$ \\ 
  \textbf{Phil Blunsom}$^{1,2}$, \textbf{Andrew Markham}$^{1}$, \textbf{Niki Trigoni}$^{1}$ \\
  \{firstname.lastname\}@cs.ox.ac.uk \\
  $^{1}$Department of Computer Science, University of Oxford $^{2}$ DeepMind\\
}

\begin{document}
\maketitle
\vspace{-0.5cm}
\begin{abstract}
Inertial information processing plays a pivotal role in ego-motion awareness for mobile agents, as inertial measurements are entirely egocentric and not environment dependent. 
However, they are affected greatly by changes in sensor placement/orientation or motion dynamics, and it is infeasible to collect labelled data from every domain.
To overcome the challenges of domain adaptation on long sensory sequences, we propose a novel framework that extracts domain-invariant features of raw sequences from arbitrary domains, and transforms to new domains without any paired data. 
Through the experiments, we demonstrate that it is able to efficiently and effectively convert the raw sequence from a new unlabelled target domain into an accurate inertial trajectory, benefiting from the physical motion knowledge transferred from the labelled source domain.
We also conduct real-world experiments to show our framework can reconstruct physically meaningful trajectories from raw IMU measurements obtained with a standard mobile phone in various attachments. 
\end{abstract}
\vspace{-0.3cm}

\section{Introduction}
\vspace{-0.2cm}
Egomotion awareness plays a vital role in developing perception, cognition, and motor control for mobile agents through their own sensory experiences \cite{Agrawal2016}.
Inertial information processing, a typical egomotion awareness process operating in the human vestibular system \cite{Cullen2012} contributes to a wide range of daily activities.
Modern micro-electro-mechanical (MEMS) inertial measurements units (IMUs) are analogously able to sense angular and linear accelerations - they are small, cheap, energy efficient and widely employed in smartphones, robots and drones. 
Unlike other commonly used sensor modalities, such as GPS, radio and vision, inertial measurements are completely egocentric and as such are far less environment dependent.
Developing accurate inertial tracking is thus of key importance for robot/pedestrian navigation and for self-motion estimation \cite{Harle2013}.

Recent work in neural inertial tracking \cite{Chen2018} has demonstrated that deep neural networks are capable of extracting high level motion representations (displacement and heading angle) from raw IMU sequence data, and providing accurate trajectories. However, the task of turning inertial measurements into pose and odometry estimates is hugely complicated by the fact that different placements (e.g. carrying a smartphone in a pocket or in the hand) and orientations lead to significantly different inertial data in the sensor frame. For example, the uncertainties of phone placements, the corresponding motion dynamics, and the projection of gravity significantly alter the inertial measurements acquired from different domains (sensor frames) while the actual trajectories in the navigation frame are identical. The data-driven method that requires substantial labelled data for training, and a model trained on a single domain-specific dataset may not generalise well to new domains. It is clearly infeasible to collect labelled data from every possible attachment, as this requires specialized motion capture systems and a high degree of effort.
In this paper, therefore, we propose a robust generative adversarial network for sequence domain transformation which is able to directly learn inertial tracking in unlabelled domains without using any paired sequences. 

We note that it is possible to train end-to-end deep neural networks when presented with large amounts of labelled data. The question becomes, how can we generalize to an arbitrary attachment in the absence of labels or a paired/time-synchronized sequence? Although from the observation the raw inertial data for each domain is very different, and the resulting odometry trajectories are also unrelated to one another, the underlying statistical distribution of odometry pose updates, if derived from a common agent (e.g. human motion), must be similar. Our intuition is to decompose the raw inertial data into a domain-invariant semantic representation, learning to discard the domain-specific motion sequence transformation. 

To overcome the challenges of generalising inertial tracking across different motion domains, we propose the \textbf{MotionTransformer} framework with Generative Adversarial Networks (GAN) for sensory sequence domain transformation. Its key novelty is in using a shared encoder to transform raw inertial sequences into a domain-invariant hidden representation, without the use of any paired data. Different from many GAN-based sequence generation models applied in the field of natural language processing \cite{Lample2018}, where the sequences consist of discrete symbols or words, our model is focused on transferring continuous long time series sensory data.

\vspace{-0.2cm}
\section{Model}
\vspace{-0.2cm}
\paragraph{Inertial Tracking Physical Model}
Instead of directly predicting the trajectories conditioned on IMU outputs, we incorporate the neural model with a physical model for better inertial tracking inference. The physical model, derived from Newtonian Mechanics, integrates the angular rates of the sensor frame $\{ \mathbf{w}_{i} \}_{i=1}^N$ ($\mathbf{w}_i \in \mathbb{R}^3$ and $N$ is the length of the whole sequence) measured by the three-axis gyroscope into orientation attitudes. 
While the linear accelerations of the sensor frame $\{ \mathbf{a}_{i} \}_{i=1}^N (\mathbf{a}_i \in \mathbb{R}^3)$ measured by the three-axis accelerometer are transformed to the navigation frame and doubly integrated to give the position displacement, which discards the impact of the constant acceleration of gravity.
This physical model is hard to implement directly on low-cost IMUs, because even a small measurement error will be exaggerated exponentially through the integration. Recent deep-learning based inertial tracking \cite{Chen2018} breaks the continuous integration by segmenting the sequence of inertial measurements $\{(\mathbf{a}_i, \mathbf{w}_i)\}_{i=1}^N$ into subsequences. We denote a subsequence as $\mathbf{x} = \{(\mathbf{a}_i, \mathbf{w}_i)\}_{i=1}^{n}$, whose length is $n$. By taking into subsequences as inputs, a recurrent neural network (RNN) is leveraged to periodically predict the polar vector $\mathbf{y} = (\Delta l, \Delta \psi)$, which represents the heading and location displacement:
    \begin{equation}
		(\Delta l, \Delta \psi) = \text{RNN} (\{(\mathbf{a}_i, \mathbf{w}_i)\}_{i=1}^{n})
	\end{equation}
Based on the predicted $(\Delta l, \Delta \psi)$, we are able to easily construct the trajectories. 
However, it requires a large labelled dataset to build an end-to-end inertial tracking system,
and it is infeasible to label data for every possible domain due to the motion dynamics and unpredictability of device placements.

\begin{figure}
     	\centering
         \includegraphics[width=1\textwidth]{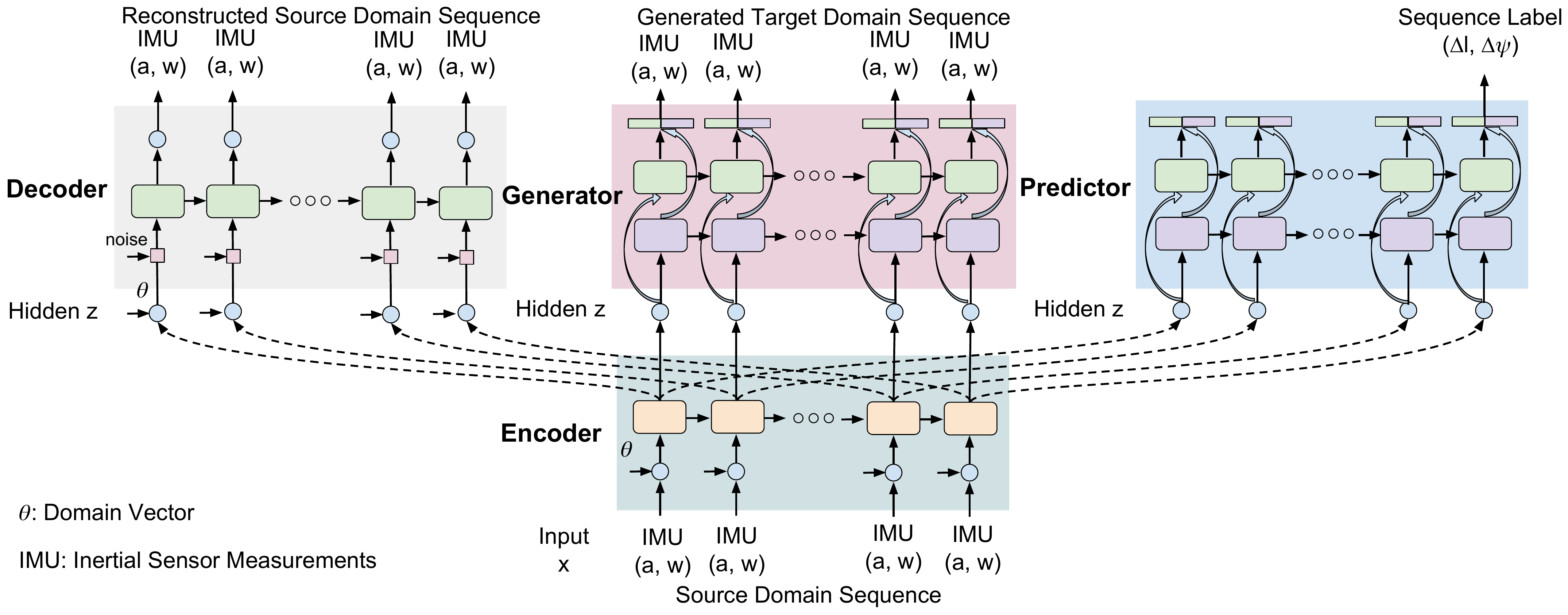}
         \caption{\label{fig:overview} Architecture of Proposed MotionTransformer: 
including the source domain sequence \textbf{Encoder} (extracting common features across different domains), the target domain sequence \textbf{Generator} (generating sensory stream in the target domain), the sequence reconstruction \textbf{Decoder} (reconstructing the sequence for learning better representations) and the polar vector \textbf{Predictor} (producing consistent trajectory for inertial navigation). 
         The GAN discriminators and the source domain Generator are omitted from this figure.
         }
    \end{figure}

\vspace{-0.2cm}
\paragraph{MotionTransformer Framework}
Here, we introduce the MotionTransformer framework, which is able to exploit the unlabelled sensory measurements, transfer the physical knowledge learned in one domain to new domains and carry out accurate inertial tracking. As Figure.~\ref{fig:overview} illustrates, our framework consists of encoder, generator, decoder and predictor modules.
Assume a scenario of two domains: a source domain and a target domain, where the source domain has labelled sequences $(\mathbf{x}^S, \mathbf{y}^S)\in\mathbb{D}^S$ ($\mathbf{y}^S$ is the sequence label - the polar vector of $\mathbf{x}^S$), and the target domain only has unlabelled sequences $\mathbf{x}^T  \in \mathbb{D}^T$. 
Note that the sequences $\mathbf{x}^S$ and $\mathbf{x}^T$ are not aligned.
The objectives of MotionTransformer Framework are three-fold: 1) extracting domain-invariant representations $\mathbf{z}$ shared across domains; 2) generating $\mathbf{\hat{x}}^T$ in the the target domain conditioned on $\mathbf{x}^S$; 3) predicting sequence labels $\mathbf{y}^T$ in the target domain.

\vspace{-0.2cm}
\paragraph{Inference}
This section introduces the learning method for jointly training the modules of our MotionTransformer, including GAN loss $\mathcal{L}_{G}$, reconstruction loss $\mathcal{L}_{AE}$, prediction loss $\mathcal{L}_{pred}$, cycle-consistency $\mathcal{L}_{cycle}$ and perceptual consistency $\mathcal{L}_{percep}$: 
	\begin{equation}
		\mathcal{L}_{total} = \mathcal{L}_{GAN} + \lambda_1 \mathcal{L}_{AE} + \lambda_2 \mathcal{L}_{pred} + \lambda_3 \mathcal{L}_{cycle} +  \lambda_4 \mathcal{L}_{percep}
	\end{equation}
where $\lambda_1$, $\lambda_2$, $\lambda_3$, and $\lambda_4$ are the hyper-parameters used as the trade-off for the optimization process.

\vspace{-0.2cm}
\section{Experiments}
\vspace{-0.2cm}
\paragraph{Inertial Tracking Dataset}

	\begin{figure*}
    	\centering
        \begin{subfigure}[t]{0.28\textwidth}
        	\includegraphics[width=\textwidth]{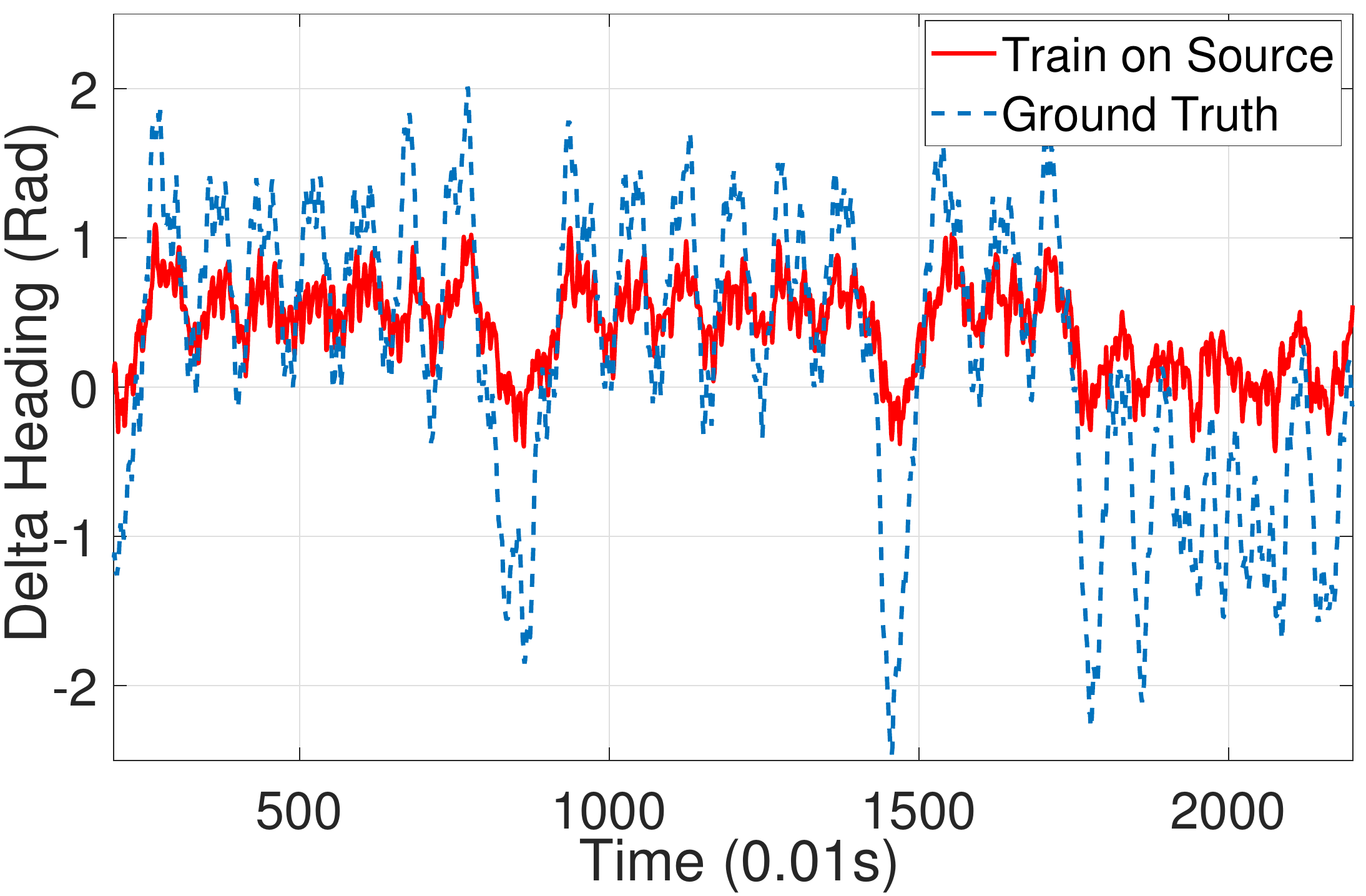}
        	\caption{\label{fig:heading source} Training in Source}
        \end{subfigure}
        \begin{subfigure}[t]{0.28\textwidth}
        	\includegraphics[width=\textwidth]{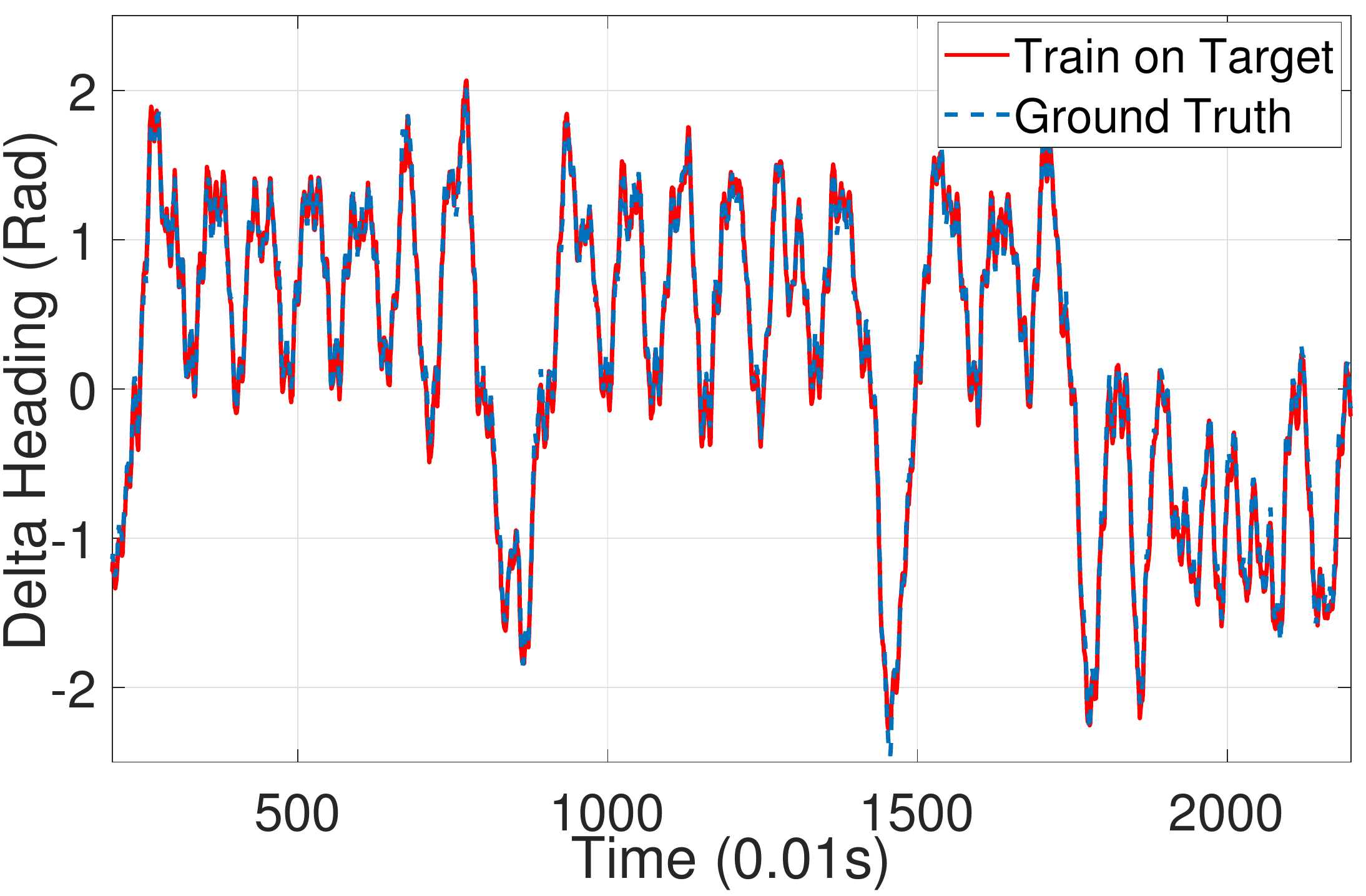}
        	\caption{\label{fig:heading target} Training in Target}
        \end{subfigure}
        \begin{subfigure}[t]{0.28\textwidth}
        	\includegraphics[width=\textwidth]{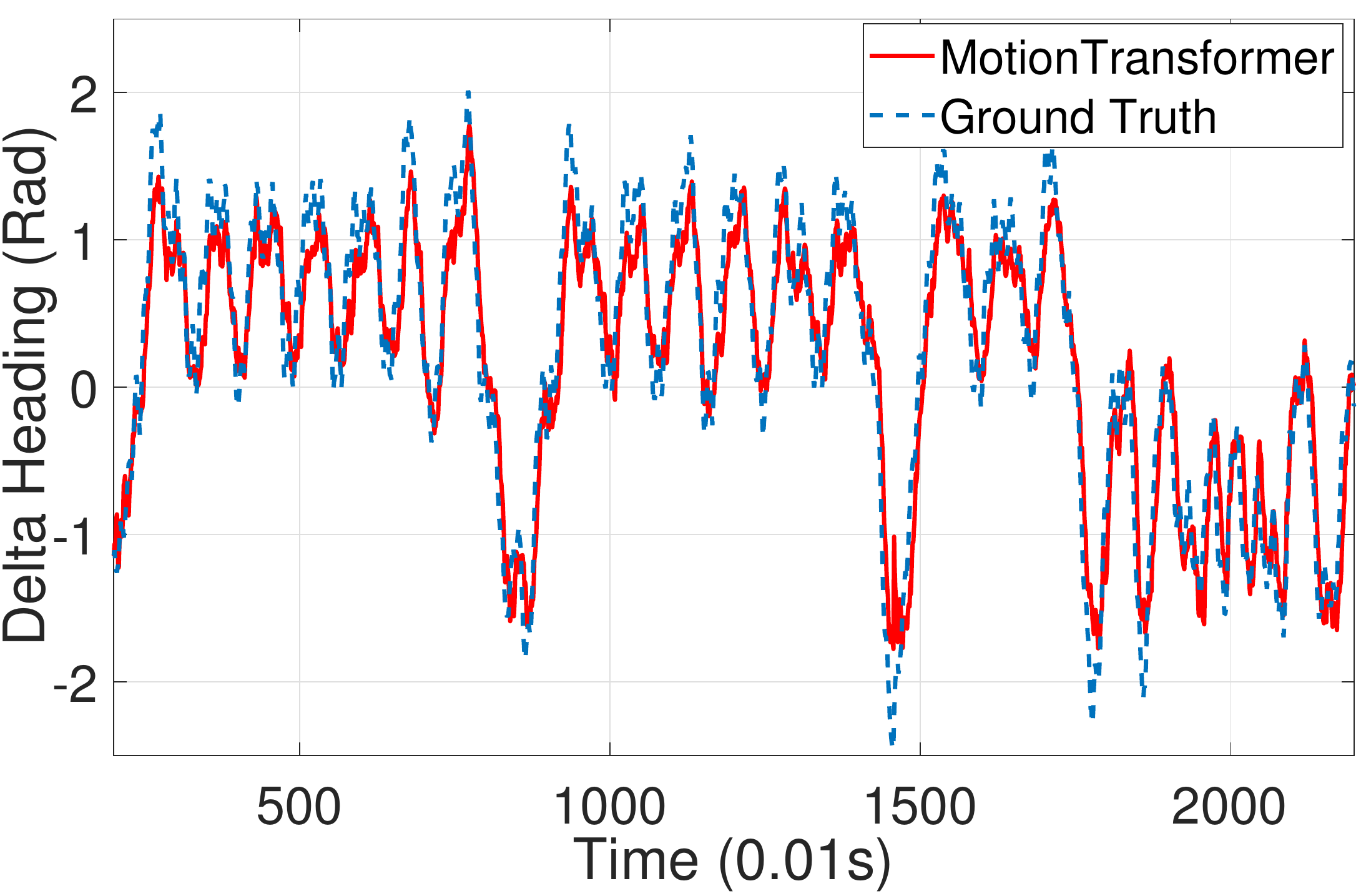}
        	\caption{\label{fig:heading da} MotionTransformer}
        \end{subfigure}
        \begin{subfigure}[t]{0.28\textwidth}
        	\includegraphics[width=\textwidth]{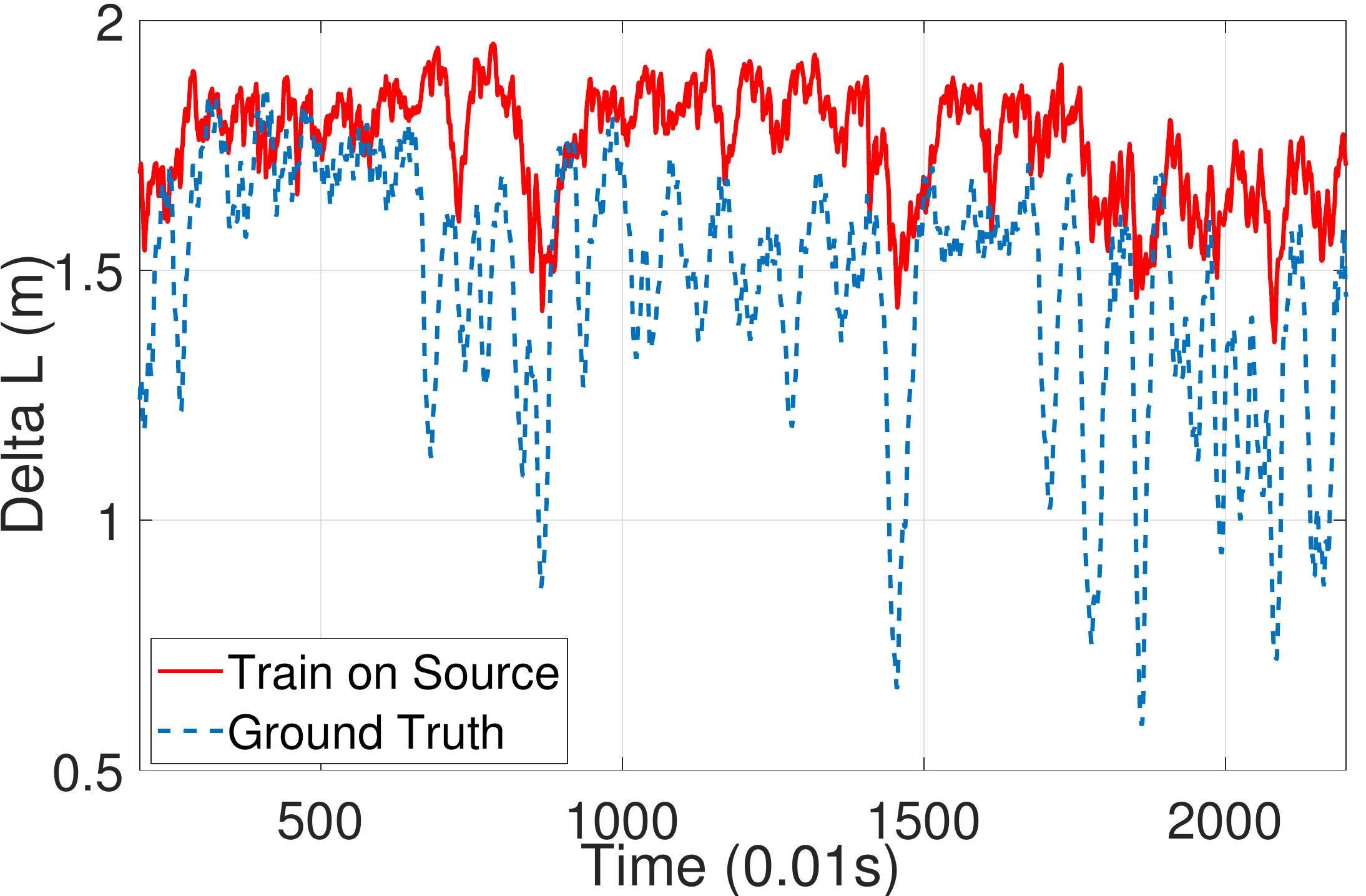}
        	\caption{\label{fig:length source} Training in Source}
        \end{subfigure}
        \begin{subfigure}[t]{0.28\textwidth}
        	\includegraphics[width=\textwidth]{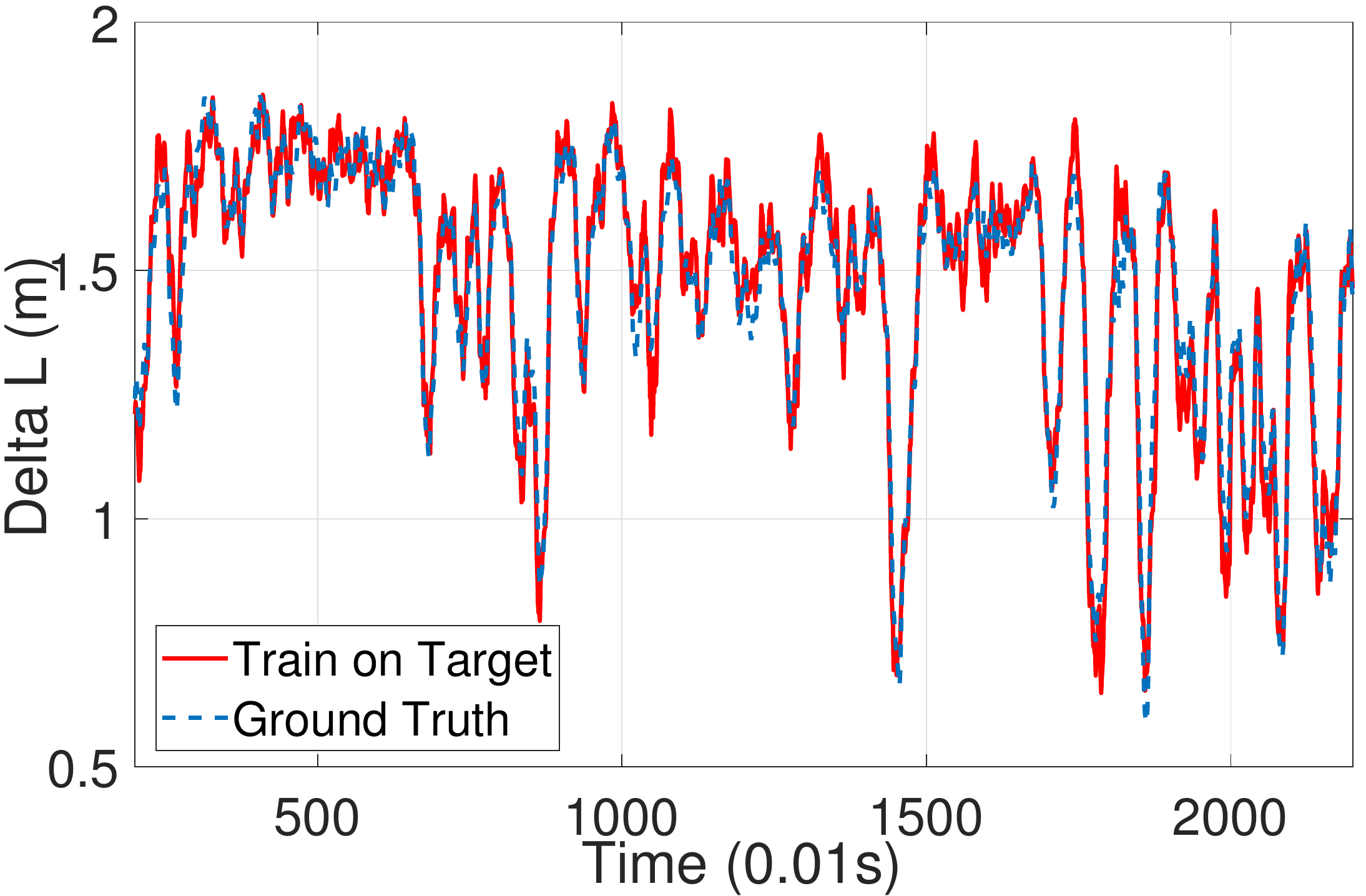}
        	\caption{\label{fig:length target} Training in Target}
        \end{subfigure}    
        \begin{subfigure}[t]{0.28\textwidth}
        	\includegraphics[width=\textwidth]{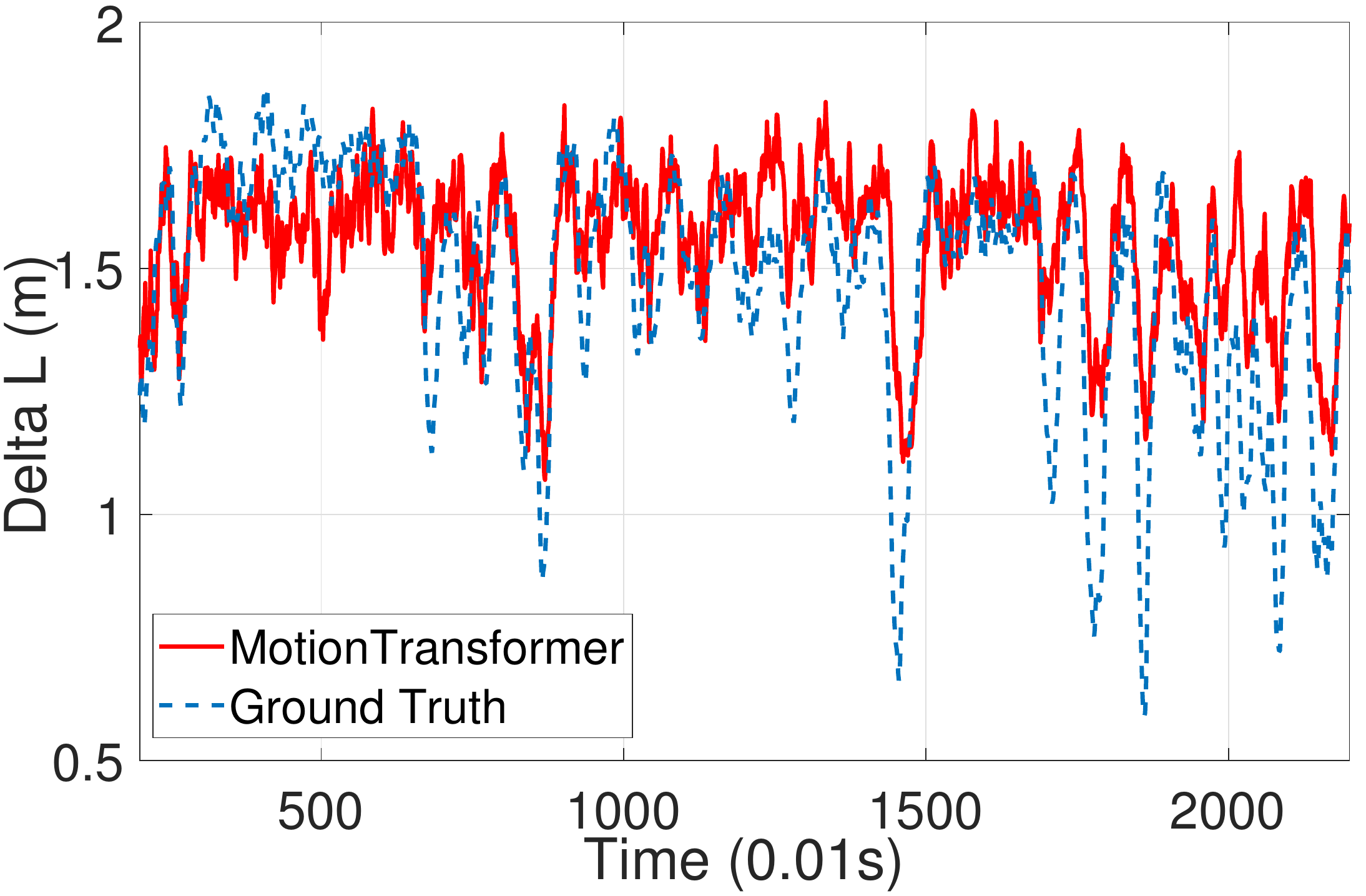}
        	\caption{\label{fig:length da} MotionTransformer}
        \end{subfigure}
        \caption{\label{fig:heading estimation} Heading displacement estimation from training in (a) source domain, (b) target domain and (c) MotionTransformer, and location displacement estimation from training in (d) source domain, (e) target domain and (f) MotionTransformer}
        \vspace{-0.5cm}
    \end{figure*}

A commercial-off-the-shelf smartphone, the iPhone 7Plus, is employed to collect inertial measurement data of pedestrian random walking \cite{chen2018oxiod} \footnote{Dataset can be found at http://deepio.cs.ox.ac.uk}. The smartphone was attached in four different poses: handheld, pocket, handbag and trolley, each of which represents a domain that has dramatically distinct motion pattern with others. We use an optical motion capture system (Vicon) to record the ground truth of motion. 
The 100 Hz sensor readings are then segmented into sequences with corresponding labels, e.g. location and heading attitude displacement provided by Vicon system. These source-domain labels are used for MotionTransformer training while the target-domain labels are used for MotionTransformer evaluation only. The length of each sequence is 200 frames (2 seconds), including three linear accelerations and three angular rates per frame. In our training phase, we use 45544, 53631, 36410 and 29001 sequences for handheld, pocket, handbag and trolley domains, and set the hyper-parameters $\lambda_1=0.01$, $\lambda_2=100$, $\lambda_3=0.1$, and $\lambda_4=1$.

\vspace{-0.2cm}
\paragraph{Transferring Across Motion Domains}
We evaluate our model on unsupervised motion domain transfer tasks. The source domain is the inertial data collected in the handheld attachment, while the target domains are those collected in the attachments of pocket, handbag and trolley. 
Its generalization performance is evaluated by comparing the label prediction (polar vector) with the ground-truth data. We compare with source-only, where we use the trained source predictor to predict data directly in the target domain and with target-only where we train the target dataset with target labels to show the performance of fully supervised learning. Figure \ref{fig:heading estimation} presents the predicted location and heading displacement in pocket domain for the three different techniques. It can be seen that source-only is unable to follow either delta heading or delta location accurately, whereas MotionTransformer achieves a level of performance close to the fully supervised target-only, especially for delta heading. 

	\begin{figure*}
    	\centering
        \begin{subfigure}[t]{0.25\textwidth}
        	\includegraphics[width=\textwidth]{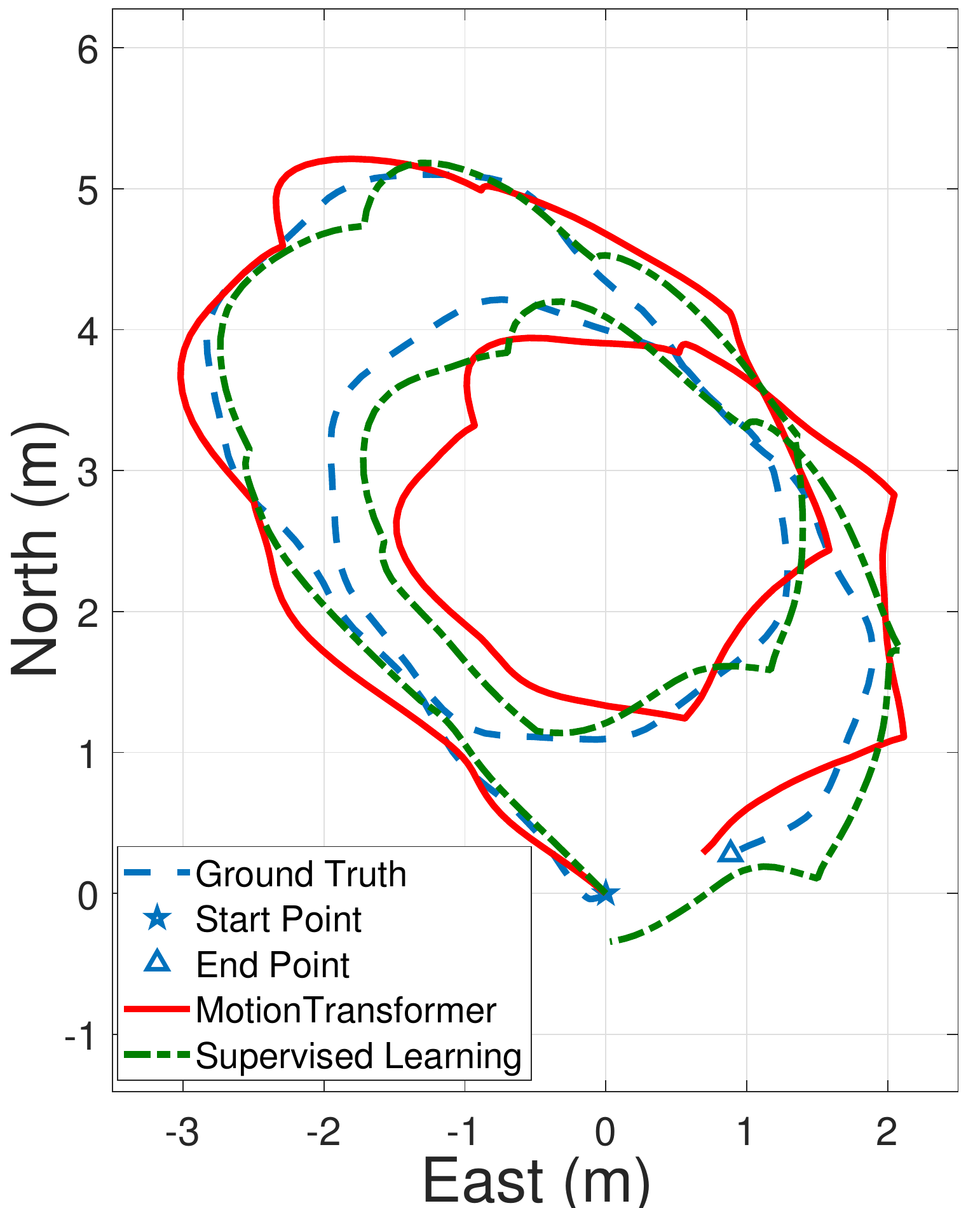}
        	\caption{\label{fig:traj_pocket} Pocket}
        \end{subfigure}
        \hspace{0.2cm}
        \begin{subfigure}[t]{0.25\textwidth}
        	\includegraphics[width=\textwidth]{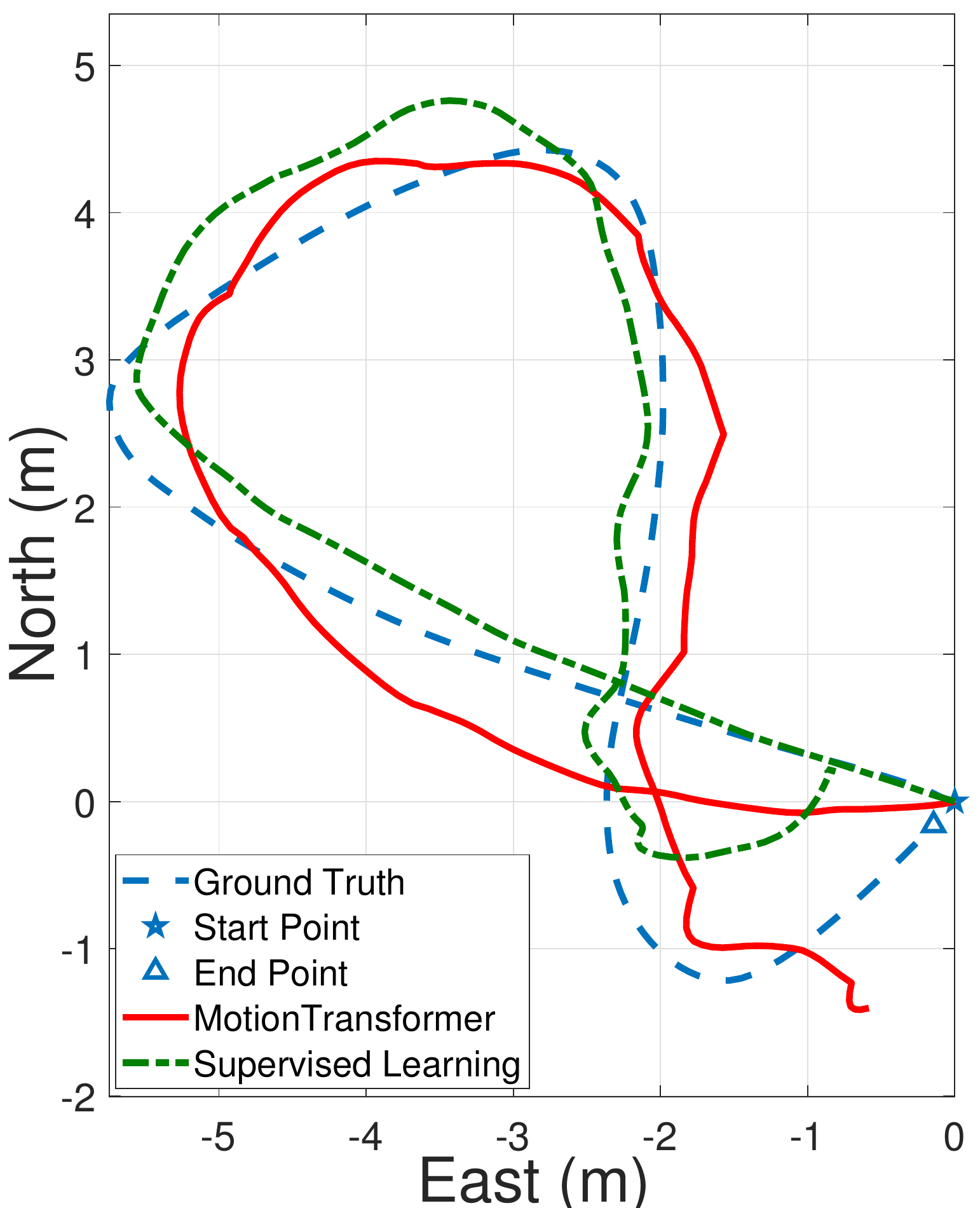}
        	\caption{\label{fig:traj_trolley} Trolley}
        \end{subfigure}
        \hspace{0.2cm}
        \begin{subfigure}[t]{0.25\textwidth}
        	\includegraphics[width=\textwidth]{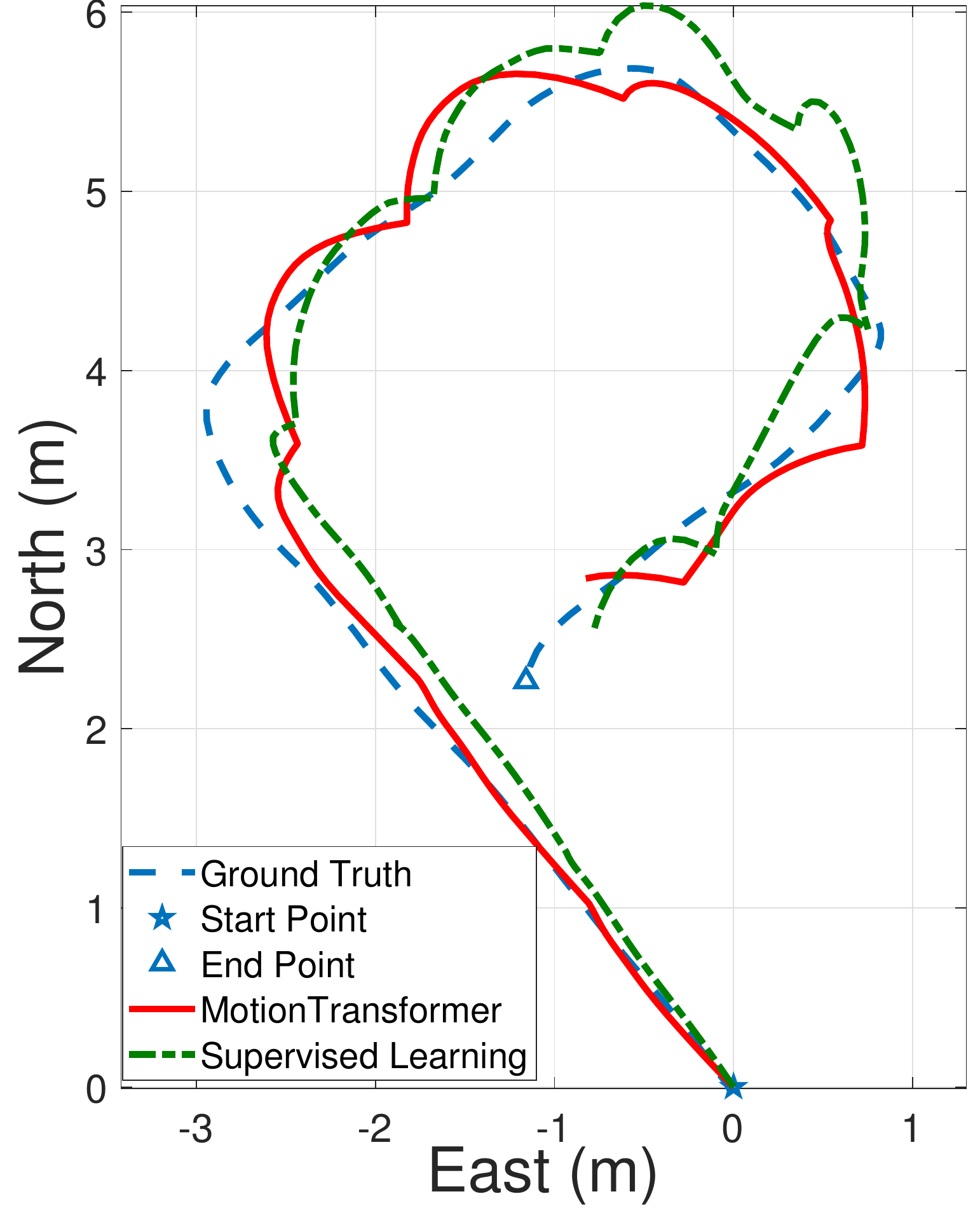}
        	\caption{\label{fig:traj_bag} Bag}
        \end{subfigure}
        \vspace{-0.2cm}
        \caption{\label{fig:traj} Inertial tracking trajectories of (a) Pocket (b) Trolley (c) Handbag, comparing our proposed unsupervised MotionTransformer with Ground Truth and Supervised Learning.
}
\vspace{-0.2cm}
    \end{figure*} 

\vspace{-0.2cm}
\paragraph{Inertial Tracking in Unlabelled Domains}
We argue that the predicted label from our domain transformation framework is capable of solving a downstream task - inertial odometry tracking. In an inertial tracking task, the precision of the predicted label determines the localization accuracy, as the current location $(x_n, y_n)$ is calculated by using an initial location $(x_0, y_0)$ and heading, and chaining the results of previous windows via Eq. \ref{eq: location update}. This dead reckoning technique, also called path integration, can be widely found in animal navigation \cite{McNaughton2006}, which enables animals to use inertial cues (e.g. steps and turns) to track themselves in the absence of vision. The errors in path integration will accumulate and cause unavoidable drifts in trajectory estimation, which imposes a requirement for accurate motion domain transformation. Without domain adaptation, if the model trained on source domain is directly applied to target domains, it will not produce any meaningful trajectory.
	\begin{equation}
    	\label{eq: location update}
    	\left\{
    	\begin{aligned}
    		x_n=x_0+\Delta l cos(\psi_0+\Delta \psi) \\
        	y_n=y_0+\Delta l sin(\psi_0+\Delta \psi)
        \end{aligned}
       \right.
    \end{equation}
We show that the inertial tracking trajectory can be recovered from the labels predicted by our domain adaptation framework in \textit{unlabelled} domains. The participant walked with the device placed in the pocket, the handbag and on the trolley. The inertial data during test walking trajectory was not included in training dataset, and collected in different days. Figure \ref{fig:traj} illustrates that our proposed model succeeds in generating physically meaningful trajectories, close to the ground truth captured by Vicon system. It proves that exploiting the raw sensory stream and transforming to a common latent distribution can extract meaningful semantic features that help solve downstream tasks.

\vspace{-0.2cm}
\section{Conclusion and Discussion}
\vspace{-0.2cm}
Motion transformation between different domains is a challenging task, which typically requires the use of labeled data for training. In the presented framework, by transforming target domains to a consistent, invariant representation, a physically meaningful trajectory can be well reconstructed. Intuitively, our technique is learning how to transform data from an arbitrary sensor domain $\theta$ to a common latent representation. Analogously, this is equivalent to learning how to translate any sensor frame to the navigation frame, without any labels in the target domain. Although MotionTransformer has been shown to work on IMU data, the broad framework is likely to be suitable for any continuous, sequential domain transformation task where there is an underlying physical model.

\vspace{-0.2cm}
\bibliographystyle{plain}
\bibliography{refs.bib}

\end{document}